\documentclass{article}
\usepackage{graphicx} 
\usepackage[a4paper, margin=1.4in]{geometry}  
\usepackage{hyperref}
\usepackage{amsmath}
\usepackage{amssymb}

\usepackage{listings}
\usepackage{xcolor}

\definecolor{codegreen}{rgb}{0,0.6,0}
\definecolor{codegray}{rgb}{0.5,0.5,0.5}
\definecolor{codepurple}{rgb}{0.58,0,0.82}
\definecolor{backcolour}{rgb}{0.95,0.95,0.92}

\lstdefinestyle{mystyle}{
    backgroundcolor=\color{backcolour},   
    commentstyle=\color{codegreen},
    keywordstyle=\color{magenta},
    numberstyle=\tiny\color{codegray},
    stringstyle=\color{codepurple},
    basicstyle=\ttfamily\footnotesize,
    breakatwhitespace=false,         
    breaklines=true,                 
    captionpos=b,                    
    keepspaces=true,                 
    numbers=left,                    
    numbersep=5pt,                  
    showspaces=false,                
    showstringspaces=false,
    showtabs=false,                  
    tabsize=2
}

\title{A Note on Shumailov et al. (2024): `AI Models Collapse When Trained on Recursively Generated Data'}

\author{Ali Borji \\
aliborji@gmail.com}
\date{June 2024}

\begin{document}

\maketitle

\begin{abstract}
The study conducted by Shumailov et al. (2024) demonstrates that repeatedly training a generative model on synthetic data leads to model collapse. This finding has generated considerable interest and debate, particularly given that current models have nearly exhausted the available data. In this work, we investigate the effects of fitting a distribution (through Kernel Density Estimation, or KDE) or a model to the data, followed by repeated sampling from it. Our objective is to develop a theoretical understanding of the phenomenon observed by Shumailov et al.  Our results indicate that the outcomes reported are a statistical phenomenon and may be unavoidable.


\end{abstract}

\section{Introduction}


\begin{figure}[ht]
    \centering
    \includegraphics[width=1\linewidth]{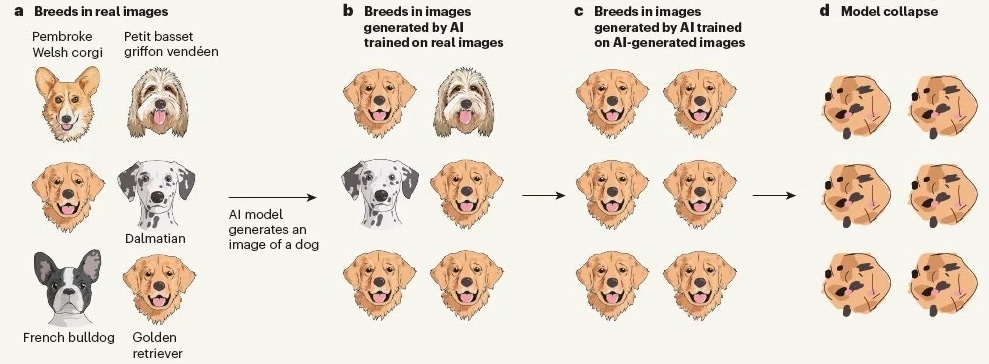}
    \caption{AI produces gibberish when trained on too much AI-generated data. Figure from ~\cite{shumailov2024ai},  \url{https://www.nature.com/articles/d41586-024-02355-z}.}
    \label{fig:teaser}
\end{figure}

The paper by Shumailov et al. 2024~\cite{shumailov2024ai} discusses how AI models collapse when trained on recursively generated data (i.e., data created by other models rather than real-world sources). It highlights that as AI-generated data proliferates, models trained on such data experience significant performance degradation (See Figure~\ref{fig:teaser}). This is due to feedback loops where models increasingly rely on lower-quality synthetic data, causing errors to compound over time. The research warns that this issue could compromise the reliability of AI systems, especially as synthetic data becomes more prevalent in training datasets.

In this work, we carry out a series of experiments to explore the generalization of these findings from a theoretical standpoint, focusing specifically on statistical sampling and distribution fitting.

\section{Experiments and results}

We seek to understand the underlying causes of this phenomenon and have considered whether a theoretical approach exists to study it systematically. The results indicate a collapse in the final distribution, with the magnitude of the effect varying according to the distance metric used between distributions. We present these findings with the hope of encouraging further exploration in this area.

\begin{figure}[t]
    \centering
    \includegraphics[width=.7\linewidth]{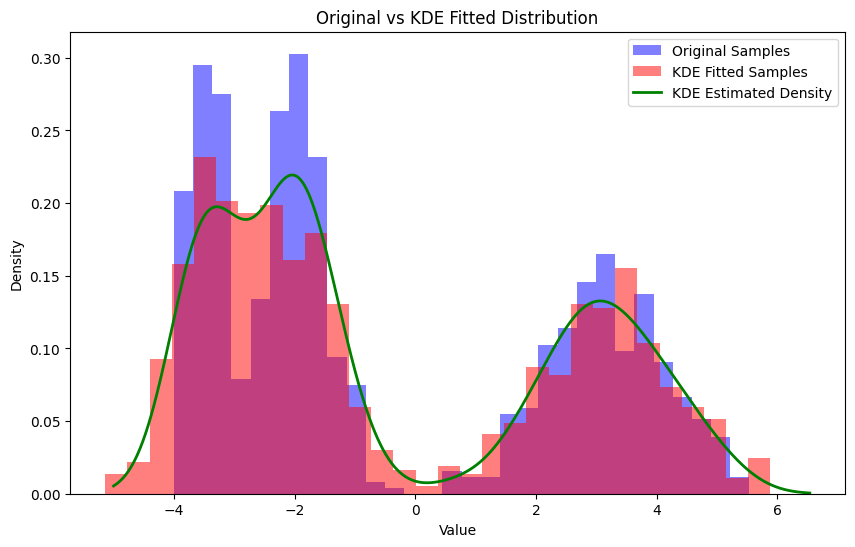}
    \caption{A synthetic distribution consisting of two Gaussian components and a Uniform distribution. The original samples and those generated using KDE are displayed. Refer to the Appendix for the code.}
    \label{fig:exp1}
\end{figure}

First, samples are drawn from a distribution composed of two normal distributions and one uniform distribution. A distribution is subsequently fitted to this data using KDE with a Gaussian kernel. Figure~\ref{fig:exp1} illustrates the data distribution, the fitted KDE, and the distribution of samples drawn from the KDE. The code in the appendix demonstrates the process.

\begin{figure}[ht]
    \centering
    \includegraphics[width=1\linewidth]{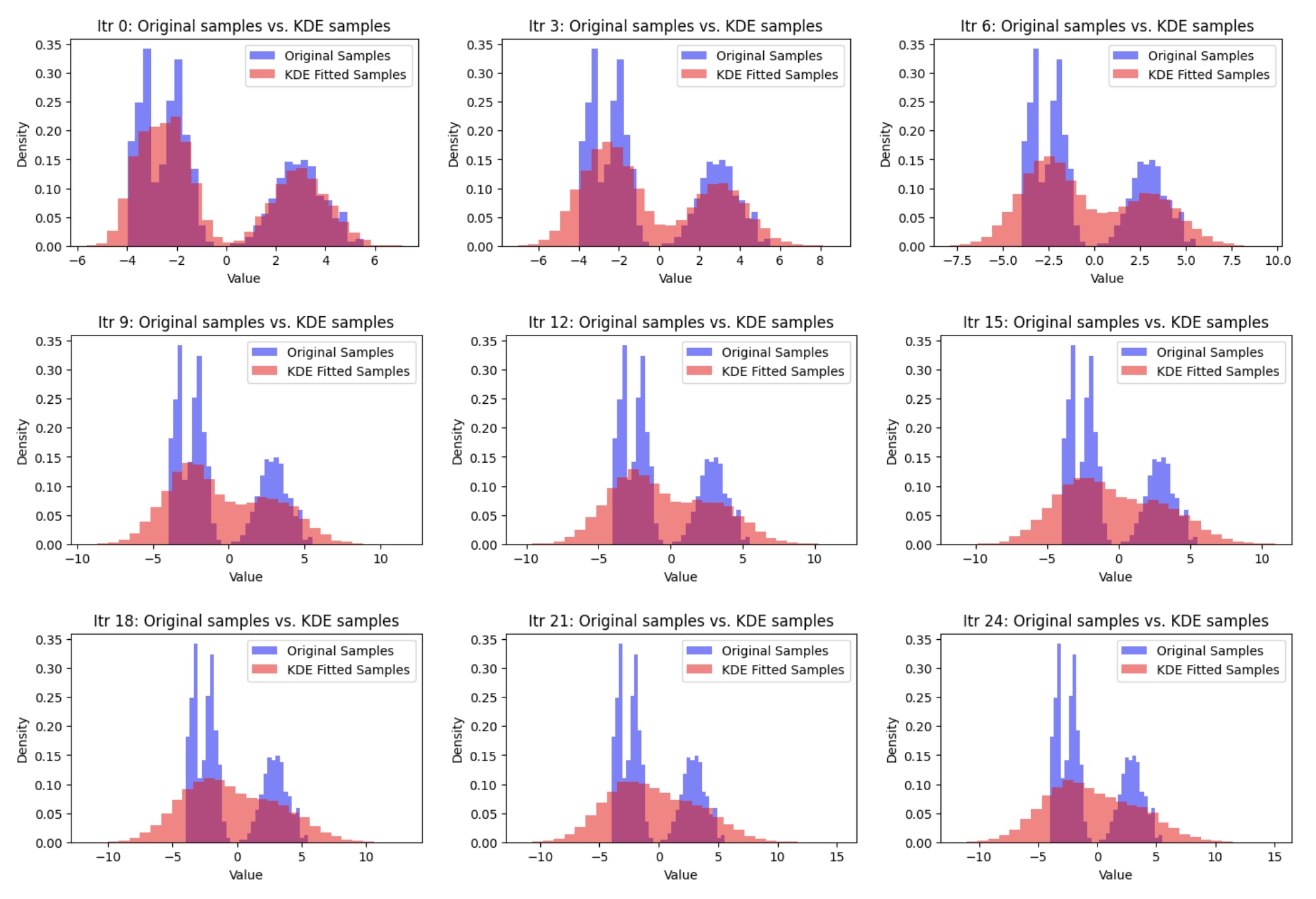}
    \caption{Recursive KDE and sampling after 30 iterations, in steps of 3, corresponding to Figure~\ref{fig:exp1}.}
    \label{fig:exp_1_res}
\end{figure}

\begin{figure}[ht]
    \centering
    \includegraphics[width=.5\linewidth]{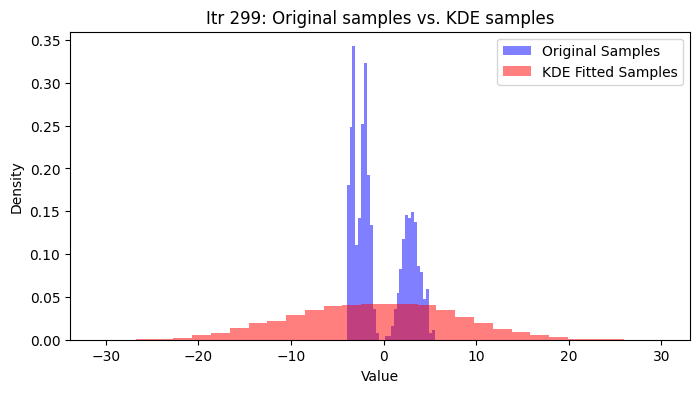}
    \includegraphics[width=.4\linewidth]{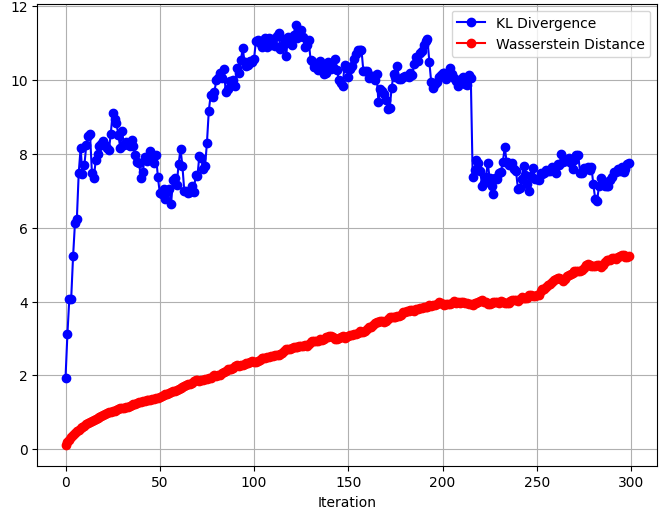}    
    \caption{Left: original distributions and samples from the KDE after 300 iterations. Right: KL divergence and Wasserstein distance over 300 iterations.}
    \label{fig:exp_1_res_cont}
\end{figure}

\begin{figure}[ht]
    \centering
    \includegraphics[width=.4\linewidth]{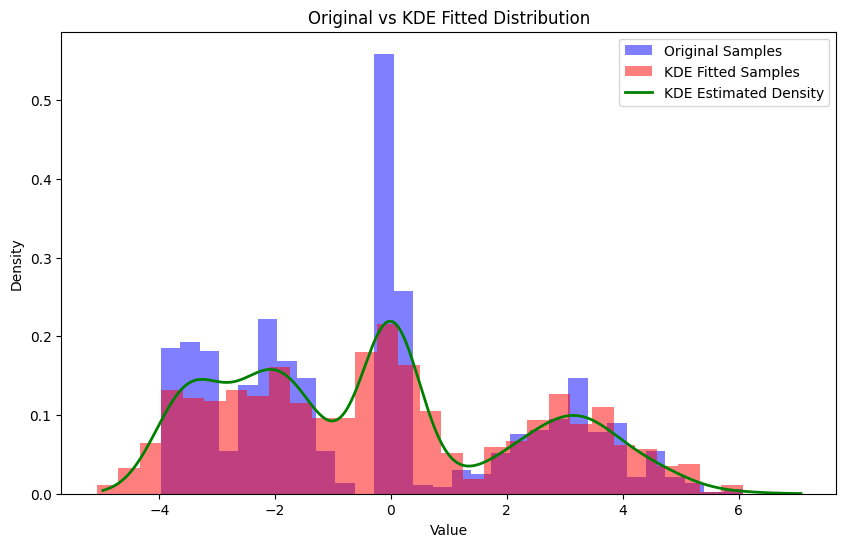} \hspace{5pt}
    \includegraphics[width=.4\linewidth]{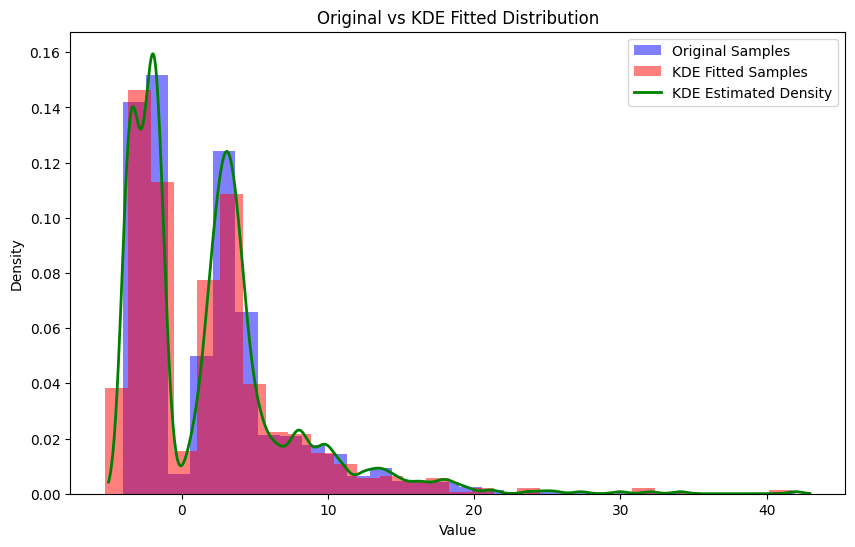}
    \includegraphics[width=.4\linewidth]{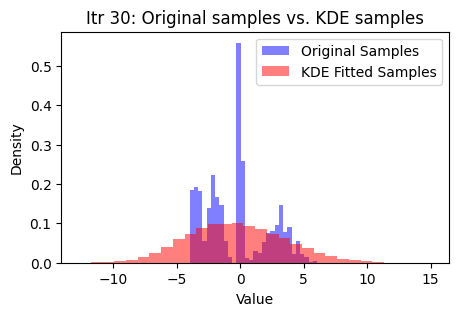} \hspace{5pt}    
    \includegraphics[width=.4\linewidth]{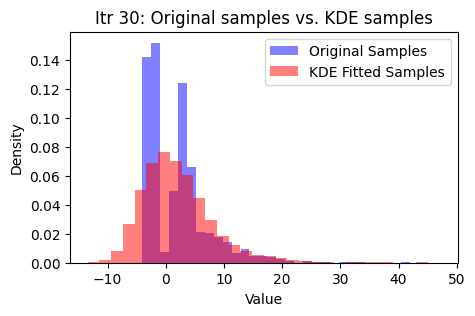}    
    \includegraphics[width=.4\linewidth]{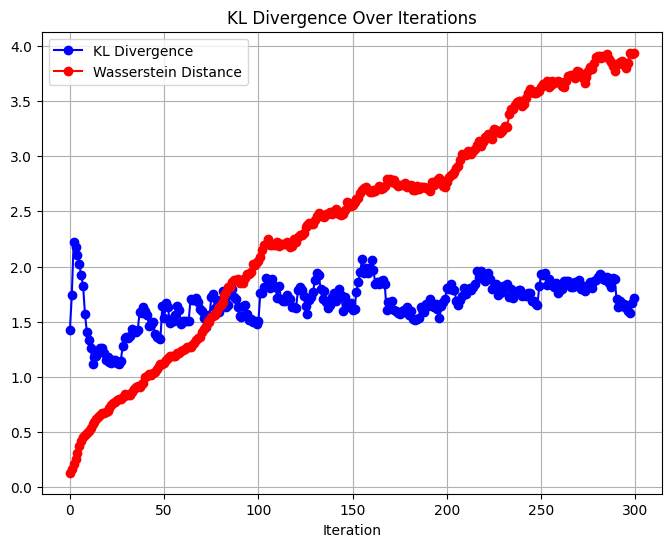} \hspace{5pt}    
    \includegraphics[width=.4\linewidth]{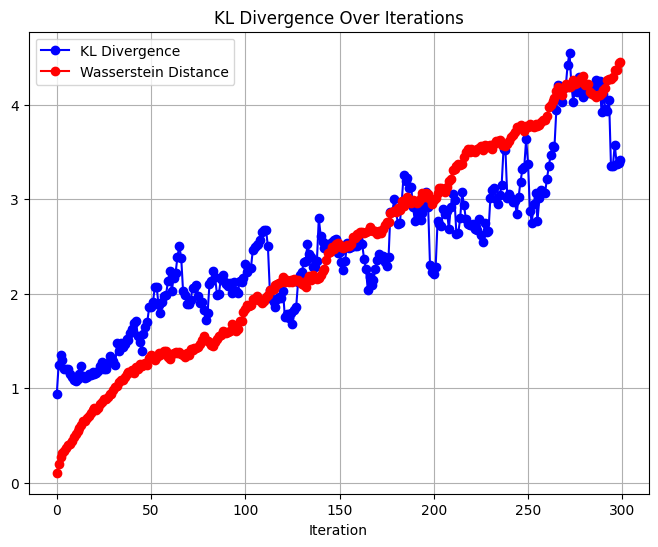}    
    
    \caption{Two additional composite distributions (one per column; see text). Top row: original distributions, fitted KDE, and samples generated from the KDE. Middle row: After 30 iterations, samples from the KDE exhibit a single mode and resemble Gaussian distributions. Bottom row: KL divergence and Wasserstein distance over 300 iterations.}
    \label{fig:exp_2_3_res}
\end{figure}

Next, we repeat the procedure and compute the KL divergence and Wasserstein distance between the data sampled from the fitted KDE and the original data.

After several iterations, the distribution progressively converges to something resembling a normal distribution, as illustrated in Figure~\ref{fig:exp_1_res}. This collapse occurs as the repeated sampling and refitting smooths out the original structure of the data, gradually losing its distinctive features and leading to a more uniform, bell-shaped curve (Figure~\ref{fig:exp_1_res}). 
With each successive iteration, both the KL divergence and Wasserstein distance (WSD) progressively increase. This suggests that the fitted distribution drifts further from the original data, indicating a growing discrepancy between the two as the process continues.

We also considered two other mixture of distributions. In the first one, 
we created a distribution composed of four distributions: three Gaussians and one uniform (See the left column in Figure~\ref{fig:exp_2_3_res}). In the second one, 
we created distribution composed of a Gamma distribution, two Gaussians and one uniform (See the left column in Figure~\ref{fig:exp_2_3_res}). In both of these cases, the distributions collapse to a uni-modal Gaussian-looking distribution.

The results of these simulations reveal that the KL divergence increases during the initial iterations but then stabilizes within a certain range (in some cases, it keeps rising). In contrast, WSD continuously grows throughout the iterations. Thus, the conclusions drawn can vary depending on the choice of distance metric.

\section{Relationship to theory}
Assume that the original data follows a normal distribution $X^0 \sim \mathcal{N}(\mu,\sigma^2)$, and we have $M_0$ samples $X^0_j$ for $j = 1, \dots , M_0$. Denoting a general sample $X^i_j$ as sample $j = 1, \dots, M_i$ at generation $i$, then the next generation model is estimated using the sample mean and variance:
\begin{equation}
\mu_{i+1} = \frac{1}{M_i}\sum_j X^i_j; \quad \sigma_{i+1}^2 = \frac{1}{M_i-1}\sum _j(X^i_j-\mu_{i+1})^2
\end{equation}

Theoretical analysis in~\cite{shumailov2024ai} has calculated variance and mean of samples at generation $n$ as follows\footnote{See also \url{https://en.wikipedia.org/wiki/Model_collapse}}:

\begin{equation}
\frac{1}{\sigma^2}\operatorname{Var}(X^n_j) = \frac{1}{M_0}+\frac{1}{M_1}+ \dots + \frac{1}{M_{n-1}}+1 + \mathcal{O}\left(M_i^{-2}\right)
\end{equation}

If all sample sizes $M_i = M$ are constant, this diverges linearly as $n\to\infty$: 

\begin{equation}
\text{Var}(X^n_j) = \sigma^2\left(1+\frac{n}{M}\right); \quad \mathbb{E}(X^n_j) = \mu
\end{equation}

It is possible to compute the distance between the true distribution and the approximated distribution at step $n+1$ using the Wasserstein-2 distance:
\begin{eqnarray}
\mathbb{E}\left[\mathbb{W}^2_2\left(\mathcal{N}(\mu,\sigma^2),\mathcal{N}(\mu_{n+1},\sigma^2_{n+1})\right)\right]&=&\frac{3}{2}\sigma^2\left(\frac{1}{M_0}+\frac{1}{M_1}+ \dots + \frac{1}{M_{n}}\right)+\mathcal{O}\left(M_i^{-2}\right) \\
\text{Var}\left[\mathbb{W}^2_2\left(\mathcal{N}(\mu,\sigma^2),\mathcal{N}(\mu_{n+1},\sigma^2_{n+1})\right)\right]&=&\frac{1}{2}\sigma^4\left(\frac{3}{M_0^2}+\frac{3}{M_1^2}+ \dots + \frac{3}{M_{n}^2} + \sum_{i\neq j}\frac{4}{M_iM_j}\right)+\mathcal{O}\left(M_i^{-3}\right)
\end{eqnarray}

This directly explains why model collapse occurs in this simple model, as the distance increases with each step $n$. 

\begin{figure}[ht]
    \centering
    \includegraphics[width=.4\linewidth]{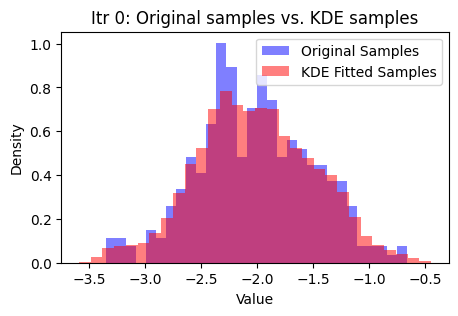} \hspace{2pt}
    \includegraphics[width=.4\linewidth]{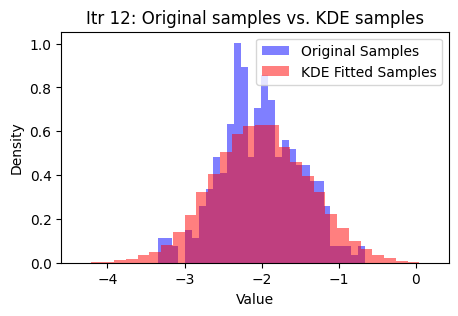} \hspace{2pt}
    \includegraphics[width=.4\linewidth]{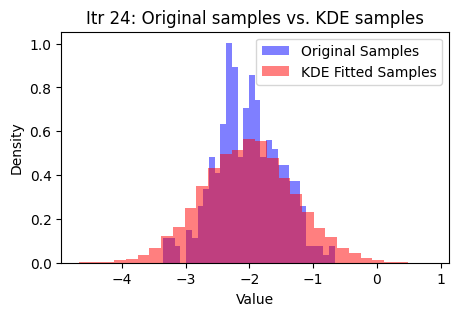} \hspace{2pt}
    \includegraphics[width=.4\linewidth]{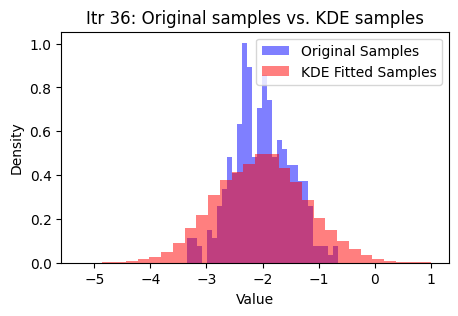} \hspace{2pt}
    \includegraphics[width=.4\linewidth]{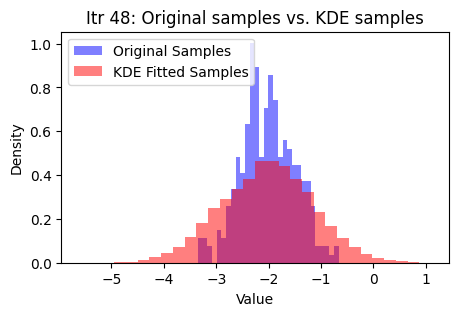} \hspace{2pt}
    \includegraphics[width=.4\linewidth]{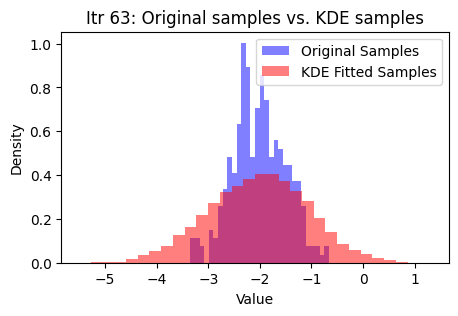} \hspace{2pt}
    \includegraphics[width=.5\linewidth]{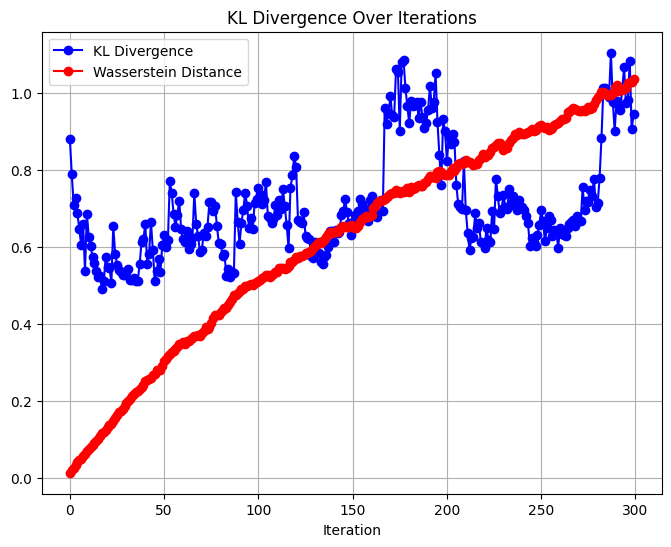} \hspace{2pt}
    
    \caption{The top panel displays a 1D Gaussian distribution along with samples from a KDE-fitted distribution after 63 iterations, taken in steps of 12. The bottom panel illustrates the distances between the two distributions over 300 iterations. Notably, while the mean of the generated samples remains consistent with the original distribution, their variance increases over time.}
    \label{fig:1D_res}
\end{figure}

The results of the simulation with a 1D Gaussian distribution are presented in Figure~\ref{fig:1D_res}. Empirically, we observe that the Wasserstein distance consistently increases with $n$ across all experiments. It is important to note that our findings hold regardless of the KDE kernel bandwidth. While a bandwidth of 0.5 was used in the previous experiments, we use a bandwidth of 0.1 here to demonstrate that the observations remain the same.

\section{Conclusion}

Overall, these findings raise important questions about whether large language models (LLMs) and other generative models are truly capable of effectively learning and capturing the underlying distributions of data. 
One of the key challenges identified is the difficulty in accurately modeling the tails of these distributions, which often contain critical information but are prone to being poorly represented by generative models. 

Measuring the distances between distributions is a particularly complex issue, as there are numerous metrics available, each with its own implications and limitations. In this paper, two such metrics are explored in detail. 

The results presented in this paper seem to reflect a broader and more pervasive phenomenon, underscoring a significant concern in the field of generative AI. Specifically, it calls attention to the potential limitations of these models in faithfully reproducing data distributions, particularly when recursively generated data is involved, which could have profound implications for their long-term reliability and robustness.

Future research in this area should investigate whether 1) our results are applicable to all types of distributions, potentially from a more theoretical perspective; 2) repeated sampling and fitting results in a uni-modal Gaussian distribution; and 3) this phenomenon is inevitable, suggesting that there may be no viable remedy.



\bibliographystyle{plain}
\bibliography{refs}

\clearpage

\section{Code}

\begin{lstlisting}[language=Python, caption=Python code for generating and plotting the mixture of distributions]
import numpy as np
import matplotlib.pyplot as plt
from sklearn.neighbors import KernelDensity

def compute_kl_divergence(p_samples, q_samples, bins=100):
    # Estimate the probability density function (PDF) using histograms
    p_hist, _ = np.histogram(p_samples, bins=bins, density=True)
    q_hist, _ = np.histogram(q_samples, bins=bins, density=True)

    # Ensure no zero values for stability
    p_hist = np.where(p_hist == 0, 1e-10, p_hist)
    q_hist = np.where(q_hist == 0, 1e-10, q_hist)

    # Calculate KL divergence
    return entropy(p_hist, q_hist)

# Generate random samples from an arbitrary distribution (e.g., mixture of Gaussians and uniform)
samples1 = np.random.normal(loc=-2, scale=0.5, size=300)
samples2 = np.random.normal(loc=3, scale=1.0, size=300)
samples3 = np.random.uniform(low=-4, high=-3, size=200)
samples = np.concatenate([samples1, samples2, samples3])

# Shuffle the samples
np.random.shuffle(samples)

# Fit a Kernel Density Estimation model to the samples
kde = KernelDensity(kernel='gaussian', bandwidth=0.5).fit(samples.reshape(-1, 1))

# Generate a range of values for plotting the KDE
x = np.linspace(min(samples) - 1, max(samples) + 1, 1000)
log_density = kde.score_samples(x.reshape(-1, 1))
density = np.exp(log_density)

# Generate new samples from the fitted KDE model
fitted_samples = kde.sample(1000).flatten()

# Plot the original samples and the fitted samples
plt.figure(figsize=(10, 6))

plt.hist(samples, bins=30, density=True, alpha=0.5, color='blue', label='Original Samples')
plt.hist(fitted_samples, bins=30, density=True, alpha=0.5, color='red', label='KDE Fitted Samples')

# Plot the KDE-estimated density
plt.plot(x, density, 'green', lw=2, label='KDE Estimated Density')

plt.title('Original vs KDE Fitted Distribution')
plt.xlabel('Value')
plt.ylabel('Density')
plt.legend(loc='best')

plt.show()
\end{lstlisting}

\begin{lstlisting}[language=Python, caption=Python code for repeated sample generation and KDE fitting]
# Parameters
num_iterations = 30  # Number of iterations
sample_size = 30000  # Number of samples to draw in each iteration

kl_divergences = []  # To store KL divergence values
wsd_divergences = []  # To store KL divergence values

# Fit the initial KDE to the original data
kde = KernelDensity(kernel='gaussian', bandwidth=0.5).fit(samples.reshape(-1, 1))

for i in range(num_iterations):
    # Sample from the current KDE model
    sampled_data = kde.sample(sample_size).flatten()

    kl_div = compute_kl_divergence(sampled_data, samples) #+ compute_kl_divergence(samples, sampled_data)
    kl_divergences.append(kl_div)

    wsd = wasserstein_distance(sampled_data, samples)
    wsd_divergences.append(wsd)

    # Print the current iteration and KL divergence
    print(f"Iteration {i + 1}: KLD = {kl_div:.4f}, WSD = {wsd:.4f}")

    # Refit the KDE model to the new sampled data
    kde = KernelDensity(kernel='gaussian', bandwidth=0.5).fit(sampled_data.reshape(-1, 1))

    if i % 3 == 0:
      # Plot the original samples and the fitted samples
      plt.figure(figsize=(5, 3))

      plt.hist(samples, bins=30, density=True, alpha=0.5, color='blue', label='Original Samples')
      plt.hist(sampled_data, bins=30, density=True, alpha=0.5, color='red', label='KDE Fitted Samples')

      # Plot the KDE-estimated density
      # plt.plot(x, density, 'green', lw=2, label='KDE Estimated Density')
      # 
      plt.title(f'Itr {i}: Original samples vs. KDE samples')
      plt.xlabel('Value')
      plt.ylabel('Density')
      plt.legend(loc='best')

      plt.show()
\end{lstlisting}

\begin{lstlisting}[language=Python, caption=Python code for generating the mixture of distributions corresponding to left column in Figure~\ref{fig:exp_2_3_res}]
# Generate random samples from an arbitrary distribution (e.g., mixture of Gaussians and uniform)
samples0 = np.random.normal(loc=0, scale=0.1, size=300)
samples1 = np.random.normal(loc=-2, scale=0.5, size=300)
samples2 = np.random.normal(loc=3, scale=1.0, size=300)
samples3 = np.random.uniform(low=-4, high=-3, size=200)
samples = np.concatenate([samples0, samples1, samples2, samples3])
\end{lstlisting}

\begin{lstlisting}[language=Python, caption=Python code for generating the mixture of distributions corresponding to right column in Figure~\ref{fig:exp_2_3_res}]
samples0 = np.random.gamma(shape=2, scale=4, size=300)
samples1 = np.random.normal(loc=-2, scale=0.5, size=300)
samples2 = np.random.normal(loc=3, scale=1.0, size=300)
samples3 = np.random.uniform(low=-4, high=-3, size=200)
samples = np.concatenate([samples0, samples1, samples2, samples3])
\end{lstlisting}

\end{document}